\documentclass[10pt, conference]{ieeeconf}
\IEEEoverridecommandlockouts  
\usepackage{amsfonts,amsmath,amssymb,siunitx}
\usepackage{graphicx}
\usepackage[dvipsnames]{xcolor} 
\usepackage{hyperref}       
\usepackage{url}            
\usepackage{booktabs}       
\usepackage{cleveref}
\usepackage[labelfont=bf,font=footnotesize]{caption}
\usepackage[most]{tcolorbox}

\graphicspath{{..}}  

\title{{\LARGE\bf
Learning Robot Behavior Priors from Non-Robotic Synthetic Data}}
\author{
Mateusz Odrowaz-Sypniewski, Michael Amir, Jasmine Bayrooti,\\
Ajay Shankar, and Amanda Prorok%
\thanks{All authors are with the Department of Computer Science and
Technology, University of Cambridge, UK; e-mails:
\texttt{\{msao2, ma2151, jgb52, as3233, asp45\}@cst.cam.ac.uk.}}%
\thanks{This work is supported by European Research Council (ERC)
Project 949940 (gAIa). J.~Bayrooti is supported by a DeepMind
scholarship, the Cambridge Trust, and by NCCR Automation, grant
agreement 51NF40\_225155 from the Swiss National Science Foundation.
We gratefully acknowledge their support.}%
}

\begin{document}
\thispagestyle{empty}
\bstctlcite{IEEE:BSTcontrol}
\maketitle

\begin{abstract}
Real robot data, rich across tasks and morphologies, is often treated as the bottleneck for training robot learning models, including embodied foundation models that aim for zero-shot physical transfer.
In this paper, we investigate 
the degree to which such models can acquire necessary structure
using data from systems \textit{without} embodiment.
To this end, we test if inexpensive, procedurally generated synthetic data can yield useful priors before a model observes any real-world behavior.
We study this through robot action prediction, a natural pretraining objective and a direct test of whether structure learned from arbitrary synthetic tasks transfers to embodied settings.
We introduce three data generation pipelines that model mathematical abstractions of robotic control, and use them as pretraining priors for a Transformer.
We demonstrate that a model trained purely on this non-robotic data attains non-trivial zero-shot prediction performance on unseen physical robot tasks via an in-context learning mechanism, and that our specialized control priors notably improve upon generic causal priors.
We also compare against a compute-matched baseline trained on real robot data, and find that it outperforms our model but only on the datasets it was trained on.
These findings suggest that procedurally generated, mathematical abstractions can yield useful robotic priors, offering a potential complement to expensive embodied data collection. 
\end{abstract}

\section{Introduction}
\label{sec:introduction}
Can robots learn from non-robotic data? Data-driven learning underpins recent advances in robot control~\cite{rt22023arxiv, bojarski2016endendlearningselfdriving, Zhao-RSS-23}, with vision-language-action (VLA) models and generalist policies relying heavily on internet-scale video, simulation, and physical demonstrations~\cite{black2026pi0visionlanguageactionflowmodel, nvidia2025gr00tn1openfoundation}.
While diversifying training data across embodiments is a compelling research direction~\cite{open_x_embodiment_rt_x_2023, lin2024data}, collecting physical demonstrations is expensive, and simulators merely shift the burden to domain-specific engineering.
Instead, we explore whether it is possible to learn from generic, entirely unembodied sources.

Consider \textit{action prediction}, the task of predicting held-out actions from a context of state-action tuples.
Intuitively, the prediction task across different robot embodiments may share basic regularities tied to general notions of control.
Prior exposure to these generic, abstract dynamics could serve as a useful starting point for downstream generalization.
Indeed, by operating downstream of any visual encoders, such a method can be \textit{complementary} to the grounding that, say, VLAs acquire from internet-scale data.

\begin{figure}[t]
    \centering
    \includegraphics[width=0.95\linewidth]{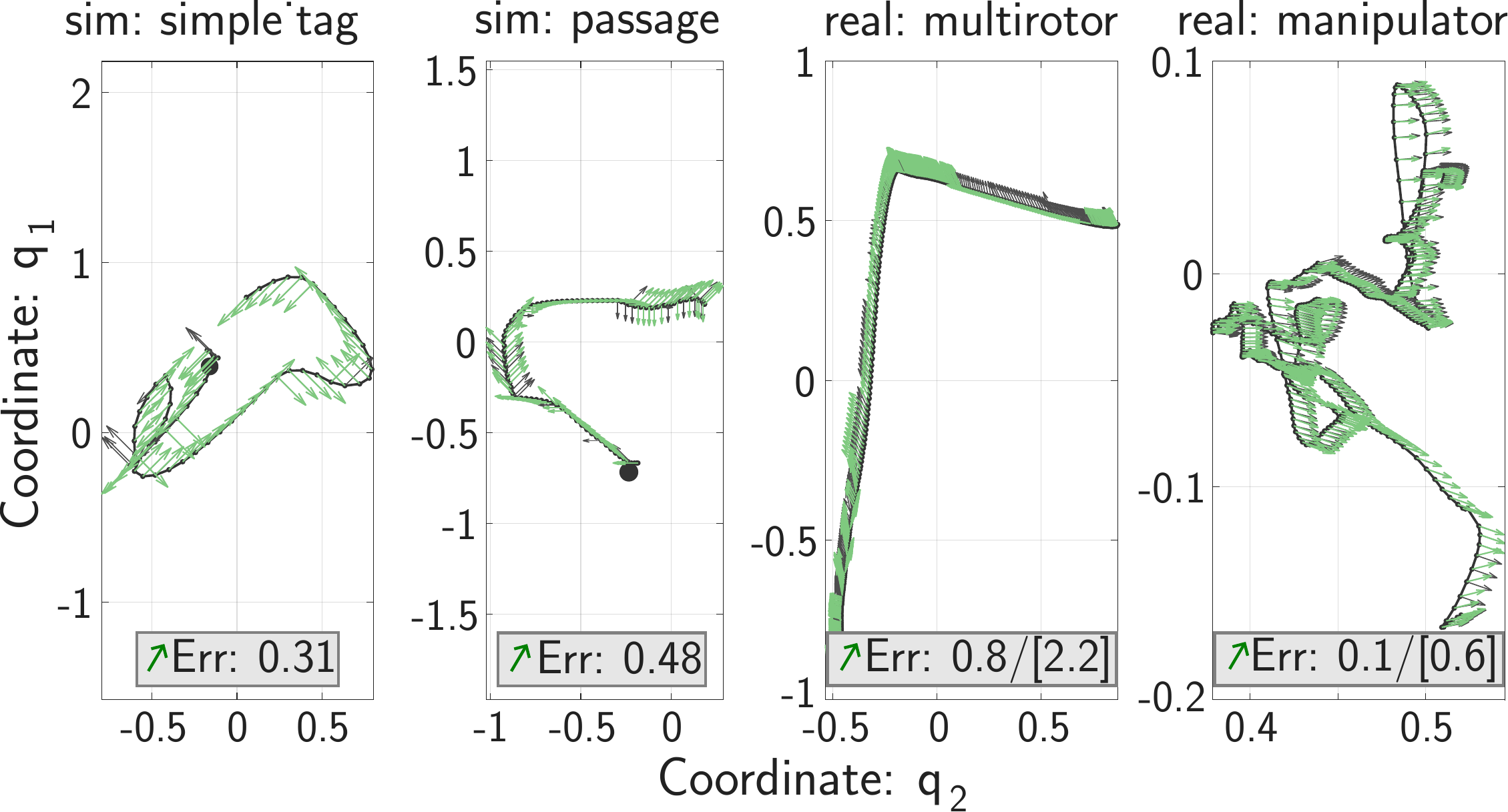}
    \caption{Predictions made by our model (\textcolor{ForestGreen}{$\rightarrow$}) and ground-truth actions ($\rightarrow$) in four sample environments.
    The model has only ever seen priors comprised of mathematical abstractions, and never any robot data (sim or real).
    Average error is shown in each panel; the simulation environments use discrete unit actions (peak norm: $\sqrt{2}$), while the peak magnitude for real environments is reported in $[\cdot]$. Only the first two coordinates $(q_1, q_2)$ are visualized.}
    \vskip-3.8ex
    \label{fig:cover}
\end{figure}

In this work, we demonstrate that generic, mathematical abstractions of control---generated through policy-induced synthetic rollouts---can induce transferable priors for robot action prediction without requiring \textit{any} embodied data.
We implement this approach using Prior-Data Fitted Networks~\cite{muller2024transformersbayesianinference}, which train Transformers on synthetic datasets sampled from task priors to enable zero-shot in-context predictions\footnote{We use `zero-shot' to denote pretraining without robot data; deployment is always an in-context (few-shot) process.}.
We hypothesize that sequential decision-making tasks require specialized structural priors, as opposed to, e.g., the domain-agnostic causal models used by methods such as TabPFN~\cite{hollmann2023tabpfn}.
Our approach to this is to develop synthetic data generators based on procedurally generated Markov Decision Processes (MDPs), Contextual Bandits, and Linear Dynamical Systems (LDS).

By utilizing only mathematical abstractions rather than physical simulations~\cite{NVIDIA_Isaac_Sim, todorov2012mujoco} or real-world data, our approach departs significantly from a typical embodied learning paradigm.
We evaluate across a diverse suite of held-out robotic datasets (\Cref{fig:cover}), and find that while a generic tabular prior struggles to transfer to real robot domains, a model pretrained purely on our synthetic control generators achieves zero-shot action prediction well above chance, despite having never observed a robot during training.
We also evaluate our approach against a compute-matched baseline trained on real, embodied robot data.
The two behave differently across the split: embodied training is far stronger on the tasks it was trained on (a gap that measures what our abstractions do not supply), while our model, for which every task is equally unseen, performs consistently across both groups and overtakes the baseline on the unseen task set.
{
\begin{tcolorbox}[colback=gray!10!white,colframe=black!25,
boxsep=1pt,before skip=7pt,after skip=0pt,left=5pt,right=4pt,
fontupper=\linespread{0.9}\selectfont]
{\small
    Our \textbf{key contribution} is in demonstrating that inexpensive synthetic data generated using purely mathematical abstractions (no sim or real data) can pretrain in-context models---enough to achieve non-trivial accuracy on fully unseen tasks.
    This can serve as a useful complement to, and is not a replacement for, training/fine-tuning using in-task data when available.
}
\end{tcolorbox}
}
\section{Related Work}
\label{sec:relatedwork}

\par\textbf{Behavioral cloning and robot data collection.}
Behavioral cloning (BC) is an approach for supervised learning from expert demonstrations,
by learning a state-to-action mapping from ground-truth state-action pairs.
BC has been applied across diverse robotic domains~\cite{10.1109/ICRA.2018.8461249, florence_manuelli_tedrake_2020, pmlr-v164-jang22a, Loquercio_2018, 2969735.2969771, bojarski2016endendlearningselfdriving}. Recent work has scaled this paradigm by training on larger datasets~\cite{open_x_embodiment_rt_x_2023}, adopting expressive architectures such as Transformers~\cite{pmlr-v155-dasari21a} and diffusion policies~\cite{10.1177/02783649241273668}, building policies on pretrained vision-language models~\cite{rt22023arxiv}, and action chunking~\cite{Zhao-RSS-23}. These approaches remain dependent on large, diverse robot datasets; insufficient scale or diversity has been shown to degrade performance~\cite{rt12022arxiv, lin2024data}.

\par\textbf{Simulators and sim-to-real robot learning.}
A popular approach to reducing the dependence on expensive real robot data is to use software that simulates agentic environments~\cite{bettini2022vmas, towers2024gymnasium} and robot physics~\cite{NVIDIA_Isaac_Sim, Mittal_Isaac_Lab_-_2025, todorov2012mujoco, jiang2024dexmimicen}. Through techniques such as GPU parallelization, these simulators offer cheap data for training and evaluating policies, and by modeling actuators and sensors they support transfer to physical robots. Our approach is complementary: rather than simulating a specific target robot or environment, we ask whether more abstract, non-robotic data can provide a useful prior for robotic tasks.

\par\textbf{Robot foundation models and cross-embodiment learning.}
Recent work has demonstrated that combining training data from different robot embodiments can improve downstream performance. Generalist policies (e.g., $\pi_0$~\cite{black2026pi0visionlanguageactionflowmodel}, GR00T~\cite{nvidia2025gr00tn1openfoundation}) train on cross-embodiment datasets, occasionally supplementing with simulation data and synthetically generated videos. Our setting differs from these works in both scale and premise: we deliberately use cheap, procedurally generated synthetic data rather than large curated robot datasets, and we ask whether our model can learn useful priors before seeing any real robot behavior.

\par\textbf{Synthetic data for in-context prediction.}
Prior-Data Fitted Networks (PFNs)~\cite{muller2024transformersbayesianinference} are trained to perform in-context learning: they take in context samples and query points from a previously unseen dataset, outputting a predictive distribution over the queries in a single forward pass. TabPFN~\cite{hollmann2023tabpfn} popularized this approach, achieving state-of-the-art performance on real tabular data despite training exclusively on synthetic structural causal models. This purely synthetic pretraining endows the model with a rich prior that adapts in-context to unseen datasets. Recent work extends this paradigm to reinforcement learning, training Transformers to predict optimal actions from policy rollouts, using either hand-specified environments~\cite{laskin2022incontextreinforcementlearningalgorithm, decisionpretrainedtransformer} or procedurally sampled ones~\cite{wang2025towards}. These works transfer within simulated decision-making; we instead ask whether priors built from non-robotic abstractions reach physical embodiments, and which abstractions matter, comparing bandits, MDPs, and linear dynamics against a generic causal prior.
\section{Problem Setting}
\label{sec:problemsetting}

We formulate robotic action prediction as an in-context classification problem~\cite{muller2024transformersbayesianinference}. A model $q_\theta$ takes in a context set
$D_{\text{ctx}} = \{(x_i, y_i)\}_{i=1}^{n}$
and query inputs
$D_{\text{query}} = \{x_j\}_{j=1}^{m}$
to compute predictive distributions
$q_\theta(y_j \mid x_j, D_{\text{ctx}})$
in a single forward pass, requiring no gradient updates.
For datasets with continuous action spaces, we tokenize actions into $K=64$ discrete bins via $K$-means~\cite{rt22023arxiv, NEURIPS2022_90d17e88}, where $y$ represents the nearest centroid index. The model minimizes a standard negative log-likelihood (NLL) objective over a dataset distribution $p_D$:
{
\begin{equation}
\small
\mathcal{L}_\theta = \mathbb{E}_{(D_{\text{ctx}}, D_{\text{query}}) \sim p_D} \left[ - \frac{1}{m} \sum_{j=1}^{m} \log q_\theta(y_j \mid x_j, D_{\text{ctx}}) \right]
\end{equation}
}

We refer to a sampled pair $(D_{\text{ctx}}, D_{\text{query}}) \sim p_D$ as a \emph{task}: a single dataset drawn from a prior, split into a context set and a set of held-out queries.

We evaluate performance using \textbf{Accuracy} and \textbf{NLL}.
To measure how well the model recovers the underlying continuous action space from discrete labels, we introduce an additional \textbf{Relative Centroid Distance (RCD)} metric.
Letting $\hat{y}_j = \arg\max_{y} q_\theta(y \mid x_j, D_{\text{ctx}})$,
RCD measures the Euclidean distance between predicted and true centroids ($c(\hat{y}_j)\  \text{and}\ c(y_j)$), normalized by the expected distance under uniform random guessing, averaged over queries:
{
\begin{equation}
\small
\text{RCD} = \mathbb{E}_{(D_{\text{ctx}}, D_{\text{query}}) \sim p_D} \left[ \frac{1}{m} \sum_{j=1}^{m} \frac{\| c(\hat{y}_j) - c(y_j) \|_2}{\frac{1}{K} \sum_{k=1}^{K} \| c(k) - c(y_j) \|_2} \right].
\end{equation}
}
Under this metric, perfect prediction yields $0.0$, and random guessing yields $1.0$.

\section{Method}

We pretrain models on synthetic episodes drawn from a mixture of generators, producing in-context tasks in the $(D_{\text{ctx}}, D_{\text{query}})$ format (\Cref{sec:problemsetting}).
To isolate the effect of the pretraining data on downstream robotic transfer, we freeze the model architecture, compute budget, and training procedure across all models within each experiment group (implementation details in \Cref{sec:reproducibility}).
The only variable between models then is the specific set of generators, $G$, that determine the task distribution $p_D$.
For each new task, we draw a generator $g \sim \mathrm{Uniform}(G)$ and sample one dataset, z-scoring its features to ensure scale consistency.
Intuitively, a good generator should exploit some structural properties that underpin any robotic task.

\subsection{Tabular (Generic) Prior}
As a baseline, we use the generic tabular prior from Hollmann et al.~\cite{hollmann2023tabpfn}.
This generator uniformly selects an underlying mechanism (either a random structural causal model (SCM), a Bayesian neural network (BNN), or a Gaussian process (GP)) to construct feature-to-label mappings, yielding static, non-sequential datasets with no embodiment structure. We retain the original maximum of 10 classes for simulated RL experiments, but modify the generator to use quantile binning with $K=64$ classes for robotic dataset experiments, matching the $K=64$ action tokens of \Cref{sec:problemsetting}.

\subsection{Synthetic Control Priors}
We hypothesize that the task of robotic action prediction rests in part on structural regularities that do not inherently require a robot: actions are selected as a function of state, states evolve smoothly under those actions, and the selection is consistent with some latent objective.
Therefore, we introduce three non-robotic generators, each capturing a complementary axis of control structure, and ask whether a model that has internalized them in a purely \textit{mathematical} setting can exploit them when they recur in embodied data:

\begin{itemize}
    \item \textbf{Markov Decision Process (MDP):} temporal, policy-induced, goal-directed sequential decision-making, where the state-to-action mapping reflects long-horizon value rather than any immediately observable signal.
    \item \textbf{Contextual Bandit:} state-conditioned action selection, single-step, isolating the state-action association with the temporal structure removed entirely.
    \item \textbf{Linear Dynamical System (LDS):} linear continuous dynamics with high temporal autocorrelation, exposing the model to the smooth state evolution that is characteristic of physical systems, which the tabular MDPs lack by construction.
\end{itemize}
Together, these span a broad spectrum of fundamental control abstractions: from localized, single-step optimization (Bandits) to smooth, temporally correlated dynamics (LDS) and complex, multi-step planning (MDP).
Crucially, none of them model a kinematic chain, a sensor, or a task.
We randomly permute the feature dimensions of every sampled dataset, so no input index carries consistent meaning across tasks; the model must infer the relevant structure from the context set at inference, exactly as it must for an unseen robot. All three generators sample a variable number of discrete actions per dataset, up to a cap $A_{max}$ matching the class count of the tabular prior in the same experiment group: $A_{max}=10$ for the simulated RL experiments (VMAS) and $A_{max}=64$ for the robotic ones.

For \textbf{MDP priors}, we generate episodes from diverse, randomly sampled tabular MDPs.
Each MDP has 30--300 states randomly distributed in $[0,1]^d$ (with dimension $d \in [10, 40]$) and 4--$A_{max}$ discrete actions acting as small spatial displacements.
Reward and continuous transition dynamics are jointly sampled from a multi-output Gaussian-process prior (via random Fourier features) to ensure smooth correlation across the state-action space.
Next-states are snapped to the nearest discrete state to preserve the tabular structure.
We solve the MDPs via truncated backward value iteration, rolling out trajectories under Boltzmann policies with random temperatures to vary exploration.
Model input is the randomly permuted raw state; the label is the discrete action.

The \textbf{contextual bandit} generator isolates single-step, state-conditioned action selection, removing temporal dynamics entirely.
Datasets consist of independent state-action pairs rather than episodic rollouts.
We sample random contextual bandits with state dimension $d_s \in [4, 100]$ and 4--$A_{max}$ discrete actions, each parameterized by a feature vector ($d_a \in [4, 16]$).
A hidden reward function $f(s, a)$ is constructed as a smooth bilinear form (of random rank) over random-Fourier-feature embeddings of states and actions, ensuring similar inputs yield similar rewards.
We construct state-action pairs by sampling states i.i.d.\ and then drawing an action for each from a Boltzmann distribution over $f(s, \cdot)$ with a random temperature.
The model input concatenates and randomly permutes the state and chosen action features, with the latter perturbed by Gaussian noise so they do not identify the chosen action, and the labels are the chosen actions at held-out states.

The \textbf{LDS} generator produces continuous dynamics with high temporal autocorrelation, exposing the model to the smooth state evolution characteristic of physical systems---a structural bias absent in the tabular MDPs.
We sample random linear dynamical systems where the state (dimension $d \in [4, 100]$) evolves as
$
x_{t+1} = A x_t + b_a + \varepsilon
$
where $A$ is a diagonal matrix with decay rates near one (inducing strong autocorrelation), $b_a$ is a fixed displacement for each of the 4--$A_{max}$ discrete actions, and $\varepsilon$ is small Gaussian noise.
We roll out multiple episodes per system using a randomly sampled linear-softmax policy, providing diverse policy demonstrations over shared dynamics.
The input is the raw state; the label is the discrete action.

\section{Evaluations}
\label{sec:result}

We evaluate the models' ability to predict next actions (given some context) on a variety of simulated and real-robot data.
For each dataset, we sample 512 windows of up to 2000 timesteps.
Each of these windows is split in half, and we predict each half from the context of the other via cross-fitting.
We report the mean and standard error across three random seeds, selected at the checkpoint with highest mean accuracy over held-out data from the training distribution.

\subsection{Datasets}
\label{sec:datasets}
We evaluate across a spectrum of task domains:
(i) six simulated multi-agent RL envs (VMAS~\cite{bettini2022vmas});
(ii) five simulated manipulation tasks (RoboMimic~\cite{robomimic2021}); and
(iii) four real-world robotic datasets across manipulator, wheeled, legged, and aerial embodiments:
DROID~\cite{khazatsky2025droidlargescaleinthewildrobot},
A2D2~\cite{geyer2020a2d2},
GrandTour~\cite{frey2026grandtour}, and
a two-multicopter flight dataset used in downwash studies~\cite{smith2023so}.
Every dataset is mapped to the format of \Cref{sec:problemsetting}: states/observations become inputs $x_t$, actions become labels $y_t$ (continuous actions are discretized into $K=64$ centroids via $K$-means).
A task is a window of timesteps from one dataset, split into $D_{\text{ctx}}$ and $D_{\text{query}}$.

\begin{figure*}
    \centering
    \includegraphics[width=0.98\textwidth]{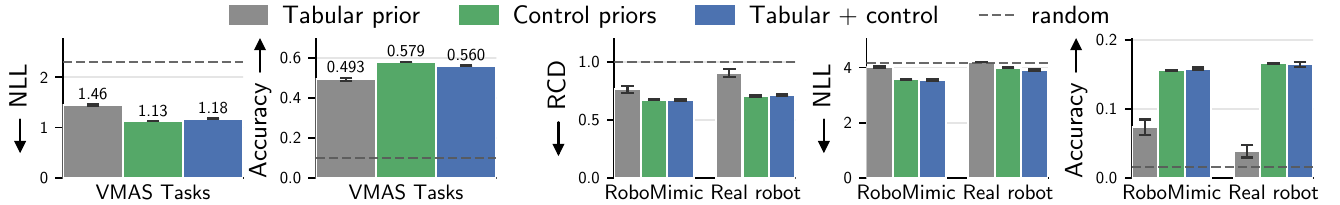}
    \caption{Performance of the model's action predictions when trained using different priors for VMAS tasks and various robot datasets. We report NLL and accuracy measures in each case, and additionally report RCD when the action space is continuous.
    A model pretrained with \textit{any} prior will already perform better than random chance (dotted line); though our control priors show a further improvement across all metrics.}
    \label{fig:vmas_robot_comparison}
\end{figure*}

\begin{table*}[!h]
\centering
\begin{minipage}[t]{0.65\linewidth}
\caption{Action prediction performance on each real-world robotic dataset for models trained using different priors, reporting mean and standard error across 3 seeds. The mean column is the grouped value plotted in \Cref{fig:vmas_robot_comparison} (Real Robot); best per column is listed in bold.}
\label{tab:per_dataset_robot}
\centering
\setlength{\tabcolsep}{5pt}
\begin{tabular}{lccccc}
\toprule
Prior & A2D2 & DROID & Aerial & GrandTour & Mean \\
\midrule
\multicolumn{6}{l}{\textit{RCD $\downarrow$}} \\
\quad Tabular prior & 0.867\,{\scriptsize$\pm$0.040} & 0.898\,{\scriptsize$\pm$0.031} & 0.923\,{\scriptsize$\pm$0.035} & 0.931\,{\scriptsize$\pm$0.088} & 0.905\,{\scriptsize$\pm$0.035} \\
\quad Control priors & 0.671\,{\scriptsize$\pm$0.030} & \textbf{0.499}\,{\scriptsize$\pm$0.004} & \textbf{0.806}\,{\scriptsize$\pm$0.003} & \textbf{0.849}\,{\scriptsize$\pm$0.010} & \textbf{0.706}\,{\scriptsize$\pm$0.008} \\
\quad Tabular + control & \textbf{0.663}\,{\scriptsize$\pm$0.028} & 0.511\,{\scriptsize$\pm$0.005} & 0.821\,{\scriptsize$\pm$0.002} & 0.859\,{\scriptsize$\pm$0.010} & 0.714\,{\scriptsize$\pm$0.005} \\
\midrule
\multicolumn{6}{l}{\textit{NLL $\downarrow$}} \\
\quad Tabular prior & 4.213\,{\scriptsize$\pm$0.025} & 4.192\,{\scriptsize$\pm$0.013} & 4.230\,{\scriptsize$\pm$0.013} & 4.144\,{\scriptsize$\pm$0.007} & 4.194\,{\scriptsize$\pm$0.003} \\
\quad Control priors & 3.805\,{\scriptsize$\pm$0.026} & 4.049\,{\scriptsize$\pm$0.007} & 4.126\,{\scriptsize$\pm$0.014} & 3.978\,{\scriptsize$\pm$0.002} & 3.990\,{\scriptsize$\pm$0.005} \\
\quad Tabular + control & \textbf{3.658}\,{\scriptsize$\pm$0.063} & \textbf{3.998}\,{\scriptsize$\pm$0.020} & \textbf{4.031}\,{\scriptsize$\pm$0.036} & \textbf{3.969}\,{\scriptsize$\pm$0.026} & \textbf{3.914}\,{\scriptsize$\pm$0.028} \\
\midrule
\multicolumn{6}{l}{\textit{Accuracy $\uparrow$}} \\
\quad Tabular prior & 0.037\,{\scriptsize$\pm$0.008} & 0.046\,{\scriptsize$\pm$0.011} & 0.038\,{\scriptsize$\pm$0.012} & 0.032\,{\scriptsize$\pm$0.004} & 0.039\,{\scriptsize$\pm$0.009} \\
\quad Control priors & \textbf{0.240}\,{\scriptsize$\pm$0.003} & \textbf{0.169}\,{\scriptsize$\pm$0.003} & \textbf{0.141}\,{\scriptsize$\pm$0.000} & 0.115\,{\scriptsize$\pm$0.001} & \textbf{0.166}\,{\scriptsize$\pm$0.000} \\
\quad Tabular + control & 0.234\,{\scriptsize$\pm$0.013} & 0.167\,{\scriptsize$\pm$0.002} & 0.136\,{\scriptsize$\pm$0.001} & \textbf{0.121}\,{\scriptsize$\pm$0.002} & 0.165\,{\scriptsize$\pm$0.004} \\
\bottomrule
\end{tabular}
\end{minipage}%
\qquad
\begin{minipage}[t]{0.3\linewidth}
\caption{Inference speed and training time for all models remain roughly consistent, with some variability due to generators.}
\label{tab:eval_timing}
\centering
\begin{tabular}{lc}
\toprule
 \textit{Inference Speed $\downarrow$}\\\textit{(context len=1000)}\\
  \begin{tabular}{l}
       Tabular prior\\Control priors\\Tabular + Control
  \end{tabular}  &  \SI{7.22}{ms/query}\\[12pt]
  \midrule
  \textit{Training Time  $\downarrow$}\\
    \begin{tabular}{l}
        Any model
    \end{tabular}& $\approx30\ \text{hours}$\\
  \bottomrule
\end{tabular}
\end{minipage}
\vskip-5pt
\end{table*}

\subsection{Evaluation of Priors}

We test three training configurations: a \textbf{Tabular Prior} (only the generic causal prior~\cite{hollmann2023tabpfn}), \textbf{Control Priors} (only our specialized synthetic priors: MDP, Contextual Bandit, LDS), and \textbf{Tabular + Control} (all four priors combined).

\textbf{VMAS Tasks.} The VMAS tasks have 9 native discrete actions (2D planar velocities), so for all priors we use the tabular prior's default class count of 10.
From \Cref{fig:vmas_robot_comparison}, we see that all models achieve \textgreater\SI{40}{\percent} accuracy on this 9-action problem (substantially higher than random chance).
The variants utilizing control priors outperform the tabular-only baseline by 7 to 9 accuracy percentage points (0.58 and 0.56 vs.\ 0.49), a margin consistent across all three seeds.
In particular, the variant trained using only control priors achieves the highest accuracy and lowest NLL, indicating that our control abstractions transfer effectively to RL tasks in simulation.

\par\textbf{Robotic Data Tasks.} For the simulated RoboMimic dataset and four real-world robot datasets (grouped together), we raise the class count of all priors to 64 to match the tokenized action spaces. 
In contrast to VMAS, performance of the models on the robotic environments diverges more visibly (\Cref{fig:vmas_robot_comparison}).
The tabular-only prior drops to within a few points of the $1/64 \approx 0.016$ chance rate on real-robot datasets, whereas mixtures incorporating control priors achieve stronger results.
While the absolute accuracy on these 64-class tasks remains modest (0.16 and 0.17 on RoboMimic and real robot data respectively for the control priors), achieving \textit{any} meaningful predictive signal without exposure to robotic training data is already a noteworthy result.
Furthermore, accuracy and NLL rankings largely align with our RCD metric, suggesting that the models capture the continuous spatial structure underlying the discrete labels. 
Ultimately, the Control generators generally match or exceed the Tabular + Control mixture (with the mixture ahead only on NLL), indicating that the control generators largely subsume the tabular prior's contribution and are not merely complementary.
We report a full breakdown of each of the priors on each dataset in \Cref{tab:per_dataset_robot}.
Inference remains real-time for all models, and their training times are also comparable since their architectures are identical (\Cref{tab:eval_timing}).

\begin{figure}
    \centering
    \includegraphics[width=0.95\linewidth]{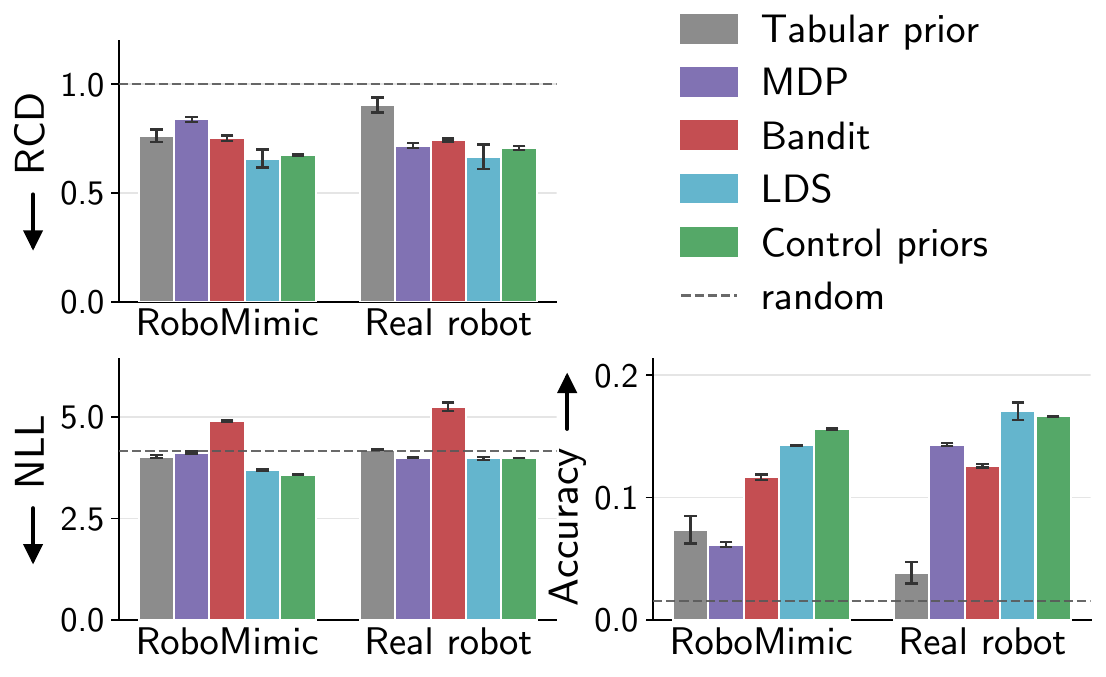}
    \caption{Ablation of the individual control priors on RoboMimic and real-world robotic datasets, alongside their combination and the tabular prior.}
    \label{fig:ablations}
\end{figure}
\subsection{Ablations}
To isolate the impact of each control prior, we train models exclusively on MDP, LDS, or Contextual Bandit. 
As shown in \Cref{fig:ablations}, our three generators do not contribute equally.
LDS is the strongest individual prior and matches the full mixture on the real-robot suite, while MDP and Bandit each fall behind the tabular baseline on at least one metric.
Note that LDS is also the only generator whose states evolve smoothly and exhibit high autocorrelation in continuous space, which suggests that what transfers is local temporal continuity in the state-action stream rather than long-horizon value structure (MDP) or state-action association alone (Bandit).
Combining all three generators still gives the best NLL and accuracy on RoboMimic, and generally remains competitive in other task sets.

\subsection{Models Trained on Embodied Data}
To ground our metrics against embodied baselines, we train two models strictly on embodied data: a \textbf{VMAS Baseline} trained on three VMAS environments (\textit{navigation, balance, reverse transport}), and a \textbf{Robotic Baseline} trained on three RoboMimic tasks (\textit{can, lift, square}) and two real-world datasets (\textit{A2D2, DROID}).
This way, each baseline is trained on a subset of the evaluation suite, with the rest held out.
This allows us to assess the value of our synthetic pretraining in a regime where in-domain embodied data is available and in one where it is not.
Our model remains zero-shot across all tasks.

\begin{figure}[t]
    \centering
    \includegraphics[width=0.95\linewidth]{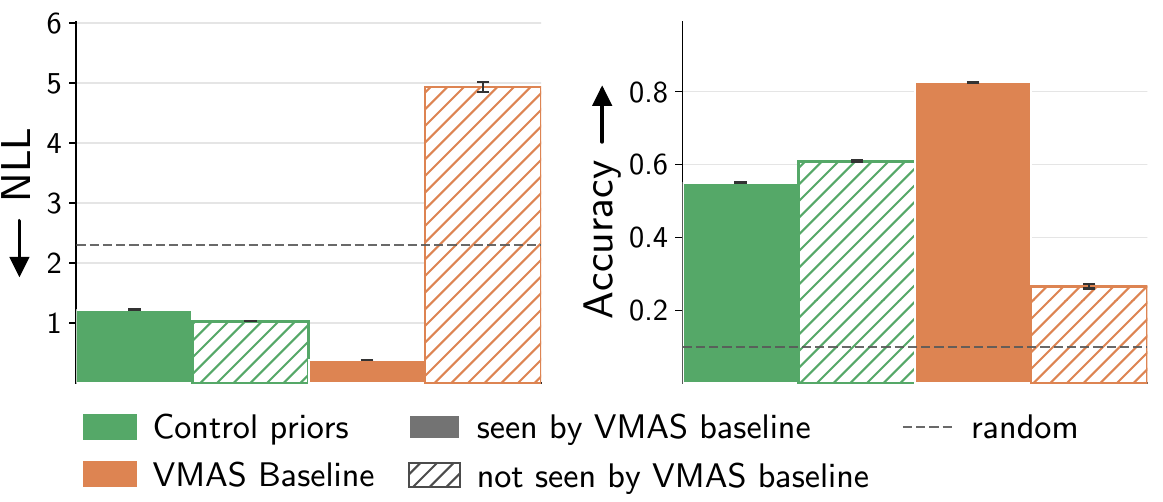}
    \caption{Comparison of synthetic pretraining (control priors) against in-domain embodied training on VMAS tasks, split by whether the baseline saw the task during training.}
    \label{fig:d_comparison_vmas}
\end{figure}

On VMAS (\Cref{fig:d_comparison_vmas}), the embodied baseline attains an accuracy of 0.83 on its training environments against our 0.55.
This gap is a clear measurement of the task-specific structure that embodied data provides but generic control priors do not.
However, the ranking is reversed on the baseline's held-out tasks (0.27 vs. 0.61); the synthetic pretraining appears to \textit{generalize} better in this case, despite the baseline having trained on environments that share dynamics and state-space characteristics with the held-out tasks.

\begin{figure}[h]
    \centering
    \includegraphics[width=\linewidth]{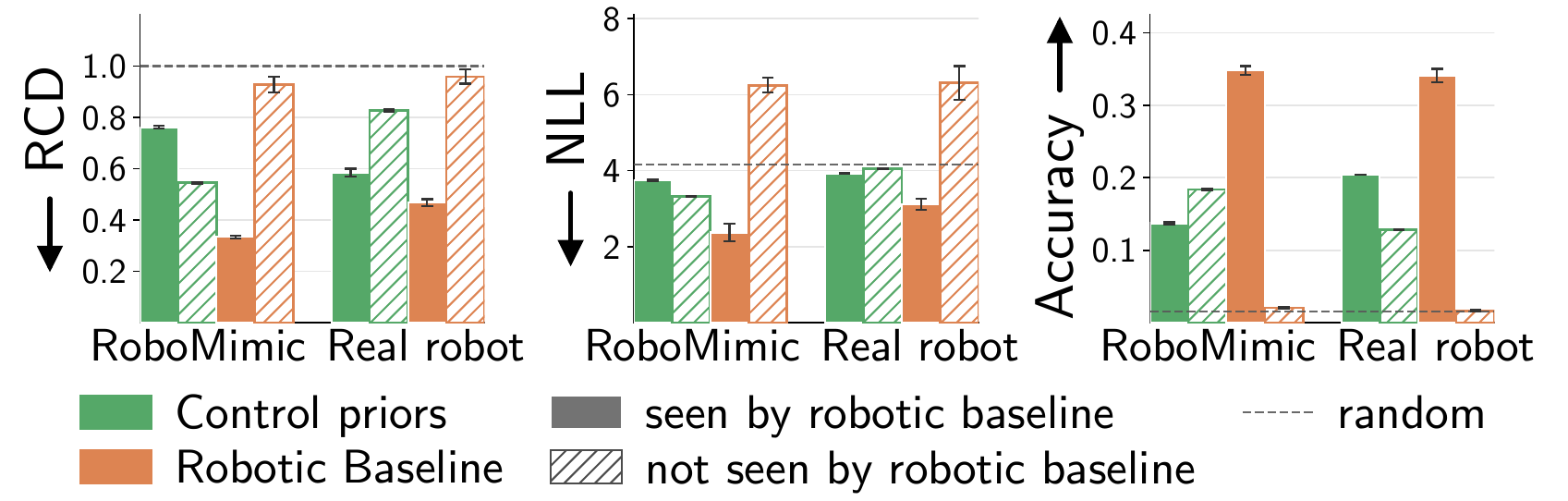}
    \caption{Comparison of synthetic pretraining against embodied training on RoboMimic and real-world robotic datasets, split by whether the baseline saw the dataset during training.}
    \label{fig:d_comparison_robotic}
\end{figure}

Results for the robotic datasets (\Cref{fig:d_comparison_robotic}) complement the simulated RL findings: the real-data baseline excels on its training datasets but falls to chance on the rest.
Meanwhile, our model, which has seen \textit{none} of these embodiments, stays well above chance throughout, and even overtakes the baseline on the datasets it did not train on.
What the baseline learns is therefore largely specific to the embodiments it saw.
We caution, however, that the severity of this collapse is not necessarily a generalizable statement about training models on (cross-)embodied data---instead, it may well reflect the limitations of our current compute budget (see scale limitations, \Cref{subsec:limitations-scale}).
Nevertheless, we remark positively upon the consistency exhibited by a model trained only using synthetic priors on seen and unseen tasks.

\subsection{Context Scaling}
\label{app:context_scaling}

\begin{figure}[t]
    \centering
    \includegraphics[width=\linewidth]{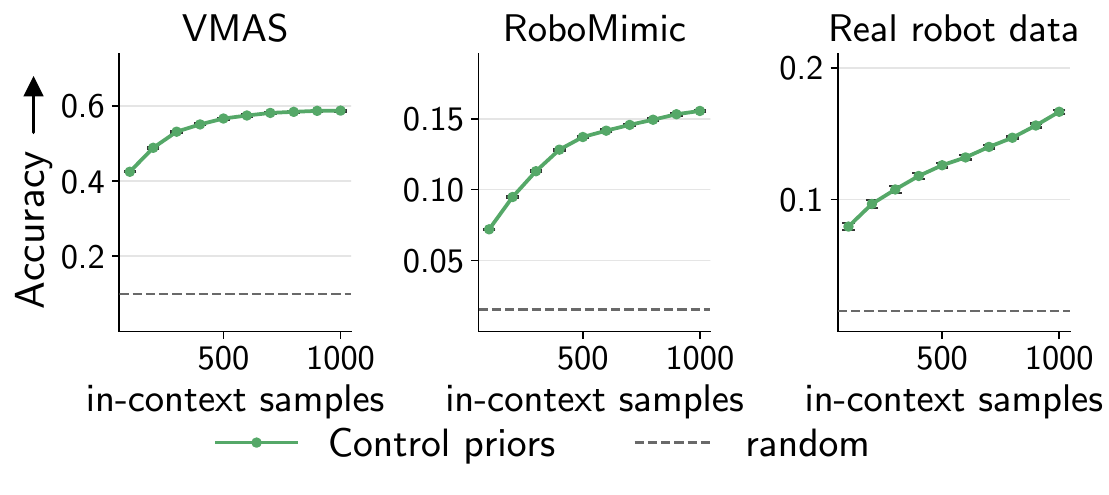}
    \captionsetup{skip=3pt}
    \caption{Accuracy of the control priors model against context size, averaged over three seeds.}
    \label{fig:context_scaling}
\end{figure}

We perform additional analyses to examine how the accuracy of our synthetic control priors model changes depending on the number of samples in the context.
\Cref{fig:context_scaling} shows the results of this experiment for context sizes $\{100, 200, \ldots, 1000\}$, with $1000$ being the context size used in our main experiments.
We observe that accuracy increases monotonically with context size in all three domains, indicating that the model extracts task-specific structure from the context rather than falling back on some context-independent decision rule.
VMAS (simulation) largely saturates by roughly 800 in-context samples, while the real-robot curve is still rising at 1000; our reported real-robot accuracies are therefore a lower bound on what the prior can support given more context.
In practice, this acts as a \textit{tradeoff:} increasing accuracy using more context is possible at the expense of compute resources and inference latency.

\section{Conclusions \& Addenda}
\label{sec:discussion}

\textit{``Can robots learn from non-robotic data?''}
We pretrained in-context models exclusively on synthetic data from mathematical abstractions (such as MDPs and LDS), and find that they exhibit non-trivial evidence of zero-shot transfer and applicability in robotic tasks.
Such models have never observed \textit{any} robot, simulated or real, and yet they are able to predict future actions reasonably in a range of eleven simulated and four real-world task sets (\Cref{fig:vmas_robot_comparison}).
Evidently, using such specialized synthetic priors offers an advantage over generic causal priors (\Cref{fig:ablations}).
We contend that these provide
an encouraging proof of concept.
Future work should certainly build on this to explore and employ this capability further.
To aid this, we present a few critical points of discussion.

The absolute accuracy of our model is relatively modest, ranging from $\approx\SI{58}{\percent}$ in simulations (where random chance would be $\approx\SI{11}{\percent}$) down to $\approx\SI{17}{\percent}$ in real-world scenarios (random chance $\approx\SI{1.6}{\percent}$).
Baselines trained on simulated or real-world data (\Cref{fig:d_comparison_vmas} and \Cref{fig:d_comparison_robotic}) invariably perform much better when they are evaluated over task sets they have seen during training.
However, we find most encouragement in cases when we evaluate on tasks that are outside the training set but share similarities in dynamics, environment, and their signals.
A model trained using only abstract synthetic priors appears to exhibit better generalization and consistency in its performance.

{
\begin{tcolorbox}[colback=gray!10!white,colframe=black!25,
boxsep=1pt,before skip=7pt,after skip=0pt,left=5pt,right=4pt,
fontupper=\linespread{0.9}\selectfont]
{\small
These results position abstract synthetic data as a scalable \textit{complement} to existing pipelines: acquiring effective structural priors may not require expensive physical experiences or embodiment-specific simulations. 
Progress toward zero-shot physical AI may therefore depend less on the volume of embodied data than current scaling trends assume.
}
\end{tcolorbox}
}

\begin{figure}[t]
    \centering
    \includegraphics[width=0.95\linewidth]{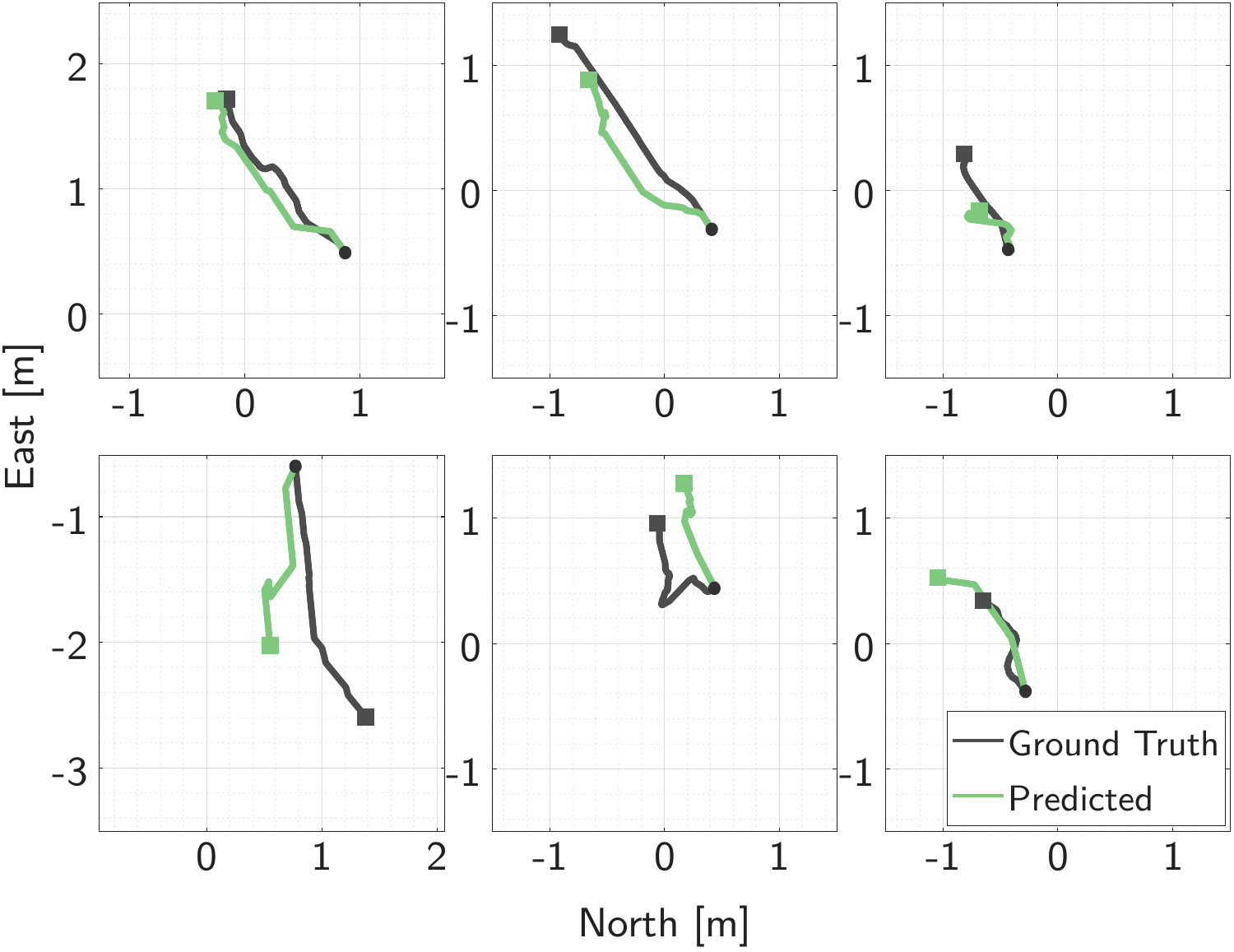}
    \caption{Six sample open-loop rollouts of 300 timesteps of ground-truth and predicted actions from a real-world multicopter flight. The solid circles and squares denote the start and the end positions, respectively.}
    \label{fig:openloop}
\end{figure}

\subsection{Open-Loop State Predictions}
Our objective of predicting future actions is useful in rolling out future states of a robot, possibly for other downstream planning and optimization tasks.
To assess the raw quality of the predictions, in \Cref{fig:openloop}, we show open-loop trajectory rollouts that na\"{i}vely use a chunk of predicted actions for 300 timesteps.
We roll out the 2D states (positions in North and East) of a multicopter using zero-order forward-step Euler integration, which is typically a poor approximation for high-order ODEs.
In addition, note that an open-loop chunk is likely to accrue substantial errors since the model does not see new (projected) states.
Nevertheless, we see that the projected final states after 300 timesteps ($\approx\SI{15}{s}$) are reasonably close to the ground-truth states (mean error $<\SI{0.48}{m}$).
The worst-case error is $\approx\SI{1}{m}$, which may be inadequate for some applications, but still suggests that the model is able to make sensible in-context predictions.

\subsection{Limitations on Scale}\label{subsec:limitations-scale}
We observed in \Cref{fig:d_comparison_robotic} that baselines trained on some robotic tasks fail to generalize to tasks outside the training set, sometimes even performing worse than random chance.
A failure in cross-embodiment transfer is not entirely surprising; however, the magnitude of the failure does warrant a qualification vis-\`{a}-vis the size/scale of our model(s).
Recent work with large models has shown noteworthy capabilities across embodiments~\cite{nvidia2025gr00tn1openfoundation, black2026pi0visionlanguageactionflowmodel}.
Our model variants (used in ablations, baselines, etc.) are substantially smaller.
Due to computational constraints, we compute-matched all baselines by employing the original, smaller version of TabPFN.
While this enables us to train models capable of real-time inference ($\approx\SI{7}{ms/query}$), their reduced capacity may explain the apparent lack of generalization.

Nevertheless, as we noted earlier, the equal-capacity models trained with synthetic priors retain their performance better.
Future work must revisit these experiments using the newer, larger TabPFN~\cite{hollmann2025tabpfn} or other models with an expanded training budget.
Since training at scale may constitute a fundamentally different learning paradigm, verifying whether the advantages of synthetic control priors persist for larger foundation models is a critical next step.

\subsection{Counter-Intuitive Fine-Tuning}
We have so far referred to our priors as ``pretraining'' the model, with the understanding that downstream tasks might utilize the capabilities in different ways.
A natural supposition is then to try to ``fine-tune'' such a model on (possibly limited) realistic, domain-specific data.
Our initial hypothesis was that with limited in-domain data, fine-tuning a model pretrained on our general control priors would yield better performance than training a new model from scratch.

To test this, we fine-tuned our pretrained model on subsets of each robotic dataset and compared its performance to a model trained from scratch on the same data.
While both models’ performance improved monotonically with subset size, their rates of improvement differed.
Surprisingly, although the pretrained model naturally outperformed the from-scratch baseline on small subsets, the latter achieved higher performance on some datasets relatively quickly, after seeing as few as 1,000 to 5,000 in-domain samples.

A detailed performance evaluation is beyond the scope here; however, these initial findings suggest that the two separate training phases may not admit a na\"{i}ve `chaining'.
This potential conflict likely stems from their fundamentally different objectives: pretraining aims for broad generalization by viewing each synthetic dataset only once, whereas fine-tuning optimizes for a given dataset through repeated exposure.
While we still believe synthetic pretraining has a place in the pipeline of foundation models, additional research is required to effectively combine it with in-domain training data.

\subsection{Datasets, Implementation, and Reproducibility}
\label{sec:reproducibility}
We will release all datasets and packaged code used for this work upon acceptance.

Table~\ref{tab:datasets} summarizes the evaluation suite. From VMAS we use six scenarios
spanning cooperative navigation (\emph{navigation}, \emph{passage}, \emph{flocking}),
pursuit (\emph{simple tag}), and collective transport (\emph{reverse transport},
\emph{balance}), rolling out pretrained policies to record (state, action) trajectories;
RoboMimic provides proficient-human demonstrations on five tasks (\emph{lift}, \emph{can},
\emph{square}, \emph{transport}, \emph{tool hang}); \emph{transport} has 127 observation
dimensions, which we truncate to the first 100 to match our model's input limit; from
DROID we use one million timesteps.
The downwash data records two multicopters in close proximity, with one subjected to the
other's aerodynamic wake, and its state stacks both vehicles' positions and velocities over
two consecutive timesteps.

\begin{table}[!t]
\centering
\caption{Evaluation datasets. $d_s$ and $d_a$ are the state and action dimensions. VMAS actions are natively discrete; all others are tokenized into $K=64$ bins via $K$-means.}
\label{tab:datasets}
\setlength{\tabcolsep}{2pt}
\begin{tabular}{@{}llccl@{}}
\toprule
Dataset & Platform & $d_s$ & $d_a$ & Action space \\
\midrule
\multicolumn{5}{l}{\textit{Simulated}} \\
\quad VMAS~\cite{bettini2022vmas} & 2D multi-agent & 19--27 & 9 & $3\times3$ motion grid \\
\quad RoboMimic~\cite{robomimic2021} & 1--2$\times$ 7-DoF Panda & 53--127 & 7--14 & $\Delta$EE pose, gripper \\
\midrule
\multicolumn{5}{l}{\textit{Real-world}} \\
\quad A2D2~\cite{geyer2020a2d2} & passenger vehicle & 9 & 3 & steer, accel., brake \\
\quad DROID~\cite{khazatsky2025droidlargescaleinthewildrobot} & 7-DoF Franka & 14 & 7 & EE target, gripper \\
\quad Aerial~\cite{smith2023so} & multicopter pair & 24 & 4 & body commands \\
\quad GrandTour~\cite{frey2026grandtour} & ANYmal quadruped & 50 & 3 & $v_x$, $v_y$, $\omega_z$ \\
\bottomrule
\end{tabular}
\end{table}

\textbf{Implementation.} We use the official implementation of Hollmann et al.~\cite{hollmann2023tabpfn}\footnote{\url{https://pypi.org/project/tabpfn/0.1.11/}}: a 12-layer Transformer with 4 attention heads. We use this earlier release rather than~\cite{hollmann2025tabpfn} both for compute reasons and because the latter does not
release its synthetic data-generation pipeline, which our method requires. We deviate from their setup only in dataset size (2048 samples, up from 1024) and training budget (15M datasets for the robotic experiments, 9M for VMAS). The embodied baselines use the same architecture and budget with the generators replaced by real data: each robotic-baseline training set is a random contiguous window of 2048 timesteps from a training split of one of its five datasets, labeled with the evaluation $K$-means codebook, while the VMAS baseline samples fresh rollouts.

\section*{Acknowledgment}
Portions of the research code were written with assistance from Claude Code. All generated code was reviewed and tested by the authors.

\bibliographystyle{IEEEtran}
\bibliography{bibctrl-ieee,example}

\end{document}